\documentclass[11pt]{article}

\usepackage[preprint]{acl}

\usepackage{times}
\usepackage{latexsym}

\usepackage[T1]{fontenc}

\usepackage[utf8]{inputenc}

\usepackage{microtype}

\usepackage{inconsolata}

\usepackage{graphicx}

\usepackage{enumitem}
\usepackage{tabularx}
\usepackage{bm,amsfonts,amsmath,dsfont}
\usepackage{subcaption}
\usepackage{multirow,multicol}
\usepackage{makecell}
\usepackage{stfloats}
\usepackage{booktabs}
\usepackage{dashrule}
\usepackage{threeparttable}
\usepackage[ruled, vlined, linesnumbered]{algorithm2e}
\usepackage{siunitx}
\usepackage{float}

\newcommand{\method}{SCALE\xspace}
\usepackage{xspace}

\makeatletter
\newcommand{\blfootnote}[1]{%
  \begingroup
  \def\@thefnmark{}%
  \@footnotetext{#1}%
  \endgroup
}
\makeatother

\title{Emotion–Cause Pair Extraction in Conversations via Semantic Decoupling and Graph Alignment}

\author{
  Tianxiang Ma\footnotemark[1], 
  Weijie Feng\footnotemark[1]\footnotemark[2], 
  Xinyu Wang, 
  Zhiyong Cheng \\
  Hefei University of Technology \\
  \texttt{matianxiang@mail.hfut.edu.cn, wjfeng@hfut.edu.cn}
}

\begin{document}
\maketitle

\renewcommand{\thefootnote}{\fnsymbol{footnote}}
\footnotetext[1]{Equal Contribution.}
\footnotetext[2]{Corresponding Author.}

\begin{abstract}

Emotion-Cause Pair Extraction in Conversations (ECPEC) aims to identify the set of causal relations between emotion utterances and their triggering causes within a dialogue.
Most existing approaches formulate ECPEC as an independent pairwise classification task, overlooking the distinct semantics of emotion diffusion and cause explanation, and failing to capture globally consistent many-to-many conversational causality.
To address these limitations, we revisit ECPEC from a semantic perspective and seek to disentangle emotion-oriented semantics from cause-oriented semantics, mapping them into two complementary representation spaces to better capture their distinct conversational roles.
Building on this semantic decoupling, we naturally formulate ECPEC as a global alignment problem between the emotion-side and cause-side representations, and employ optimal transport to enable many-to-many and globally consistent emotion-cause matching.
Based on this perspective, we propose a unified framework \method that instantiates the above semantic decoupling and alignment principle within a shared conversational structure.
Extensive experiments on several benchmark datasets demonstrate that \method consistently achieves state-of-the-art performance.
Our codes are released at https://github.com/CoCoSphere/SCALE.

\end{abstract}


\section{Introduction}

\begin{figure}[!t]
    \centering
    \begin{subfigure}{0.48\textwidth}
        \includegraphics[width=\linewidth]{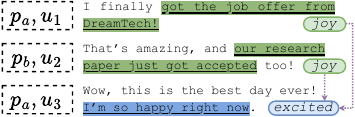}
        \caption{An example of ECPEC task.}
        \label{fig:ecpec}
    \end{subfigure}
    \vspace{-2pt}
    \noindent\hdashrule[1.2ex]{\linewidth}{0.5pt}{3pt 2pt}
    \vspace{-1pt}
    \begin{subfigure}{0.48\textwidth}
        \includegraphics[width=\linewidth]{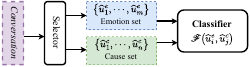}
        \caption{\textit{Select-then-pair} paradigm.}
        \label{fig:traditional}
    \end{subfigure}
    
    \vspace{-2pt}
    \noindent\hdashrule[1.2ex]{\linewidth}{0.5pt}{3pt 2pt}
    \vspace{-1pt}
    \begin{subfigure}{0.48\textwidth}
        \includegraphics[width=\linewidth]{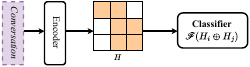}
        \caption{\textit{Embed-then-pair} paradigm.}
        \label{fig:traditional}
    \end{subfigure}
    
    \vspace{-2pt}
    \noindent\hdashrule[1.2ex]{\linewidth}{0.5pt}{3pt 2pt}
    \vspace{-1pt}
    \begin{subfigure}{0.48\textwidth}
        \includegraphics[width=\linewidth]{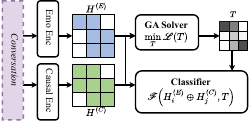}
        \caption{Our proposal.}
        \label{fig:ours}
    \end{subfigure}
    \caption{Comparison between the existing ECPEC paradigm (b-c) and \method (d).}
    \label{fig:intro}
\end{figure}

\emph{Emotion-Cause Pair Extraction in Conversations} (ECPEC) aims to identify the set of causal relations between emotion utterances and their triggering causes within a dialogue, which often exhibit complex and many-to-many dependencies.
Figure~\ref{fig:ecpec} illustrates a representative example, where the emotion utterance $u_3$ is jointly triggered by multiple preceding utterances $u_1$ and $u_2$.
Unlike emotion recognition in conversation (ERC)~\cite{wang2024a,fu2023,majumder2019}, which focuses on assigning discrete emotion labels to individual utterances~\cite{fu2021,hu2021}, ECPEC provides a causal perspective for dialogue understanding, enabling more fine-grained analysis of emotional dynamics.
As highlighted by~\citet{poria2021}, ECPEC has broad applicability across multiple domains, including dialogue systems~\citep{rashkin2019,zhong2020}, conversational recommendation~\citep{liang2024,li2018}, mental health analysis~\citep{cambria2018,pontiki2016}, and social media opinion mining~\citep{alexanderpak2010,liu2022}.


Early studies on ECPEC predominantly followed the \emph{select-then-pair} paradigm~\cite{ding2020,wang2023}, which independently identifies candidate emotion utterances and cause utterances before pairing them through heuristic or classifier-based matching, as shown in Figure~\ref{fig:traditional}.
While intuitive and easy to integrate with existing ERC models~\cite{gao2023,nguyen2024,ghosal2019}, this pipeline is prone to error propagation during candidate selection and fails to fully exploit contextualized utterance representations.
To alleviate these issues, subsequent studies shifted towards the \emph{embed-then-pair} paradigm~\cite{an2023,li2023,jeong2023,wang2024}, where utterance embeddings are directly concatenated and classified as emotion-cause pairs in an end-to-end manner. 
Although this paradigm better leverages utterance-level semantics and avoids explicit candidate construction, it still treats emotion-cause inference as a collection of independent pairwise decisions.

Despite their procedural differences, existing ECPEC approaches share two fundamental \textbf{\textit{limitations}}.
\textit{\textbf{L1)}} Most methods encode emotion-related and cause-related information within a unified representation space or interaction structure, implicitly assuming that emotion diffusion and cause explanation follow homogeneous relational patterns.
However, in real conversations, emotional states tend to propagate through contextual and speaker-dependent dynamics, whereas causes are grounded in explanatory and often asymmetric dependencies.
Conflating these distinct semantics obscures their respective roles in conversational causality.
\textit{\textbf{L2)}} Existing methods typically formulate ECPEC as independent one-to-one pair classification with binary judgments.
Such pairwise formulations are inherently inadequate for modeling globally consistent many-to-many causal structures, where multiple interdependent causes may jointly trigger an emotion and a single cause may influence multiple emotional outcomes.

To address these limitations, we revisit ECPEC from a semantic perspective and argue that emotion diffusion and cause explanation, while grounded in the same conversational structure, should be characterized by different semantic focuses.
Rather than duplicating dialogue structures or enforcing task-level separation, we seek to disentangle emotion-oriented and cause-oriented semantics by mapping them into two complementary representation spaces induced from a shared conversation graph.
Building on this semantic decoupling, we naturally formulate ECPEC as a global alignment problem between emotion-side and cause-side representations, which enables holistic reasoning over many-to-many emotion--cause relations.
Based on this perspective, we propose \textbf{\method} (\textbf{S}emantic \textbf{C}ausal \textbf{AL}ignment for \textbf{E}CPEC), a unified framework that instantiates semantic decoupling and global alignment within conversational contexts.
Extensive experiments on multiple benchmark datasets demonstrate that \method consistently outperforms existing state-of-the-art approaches.
Overall, the main contributions of this work are summarized as follows:
\begin{itemize}
    \item We revisit ECPEC from a semantic perspective and highlight the necessity of disentangling emotion diffusion and cause explanation while preserving shared conversational structure.
    \item We propose \method, a unified framework that induces emotion-side and cause-side representations from a shared conversation graph and formulates ECPEC as a global alignment problem to support many-to-many and globally consistent inference.
    \item Extensive experiments on several public ECPEC benchmarks demonstrate that \method consistently achieves state-of-the-art performance.
\end{itemize}

\section{Methodology}

\begin{figure*}[!t]
    \centering
    \includegraphics[width=.99\textwidth]{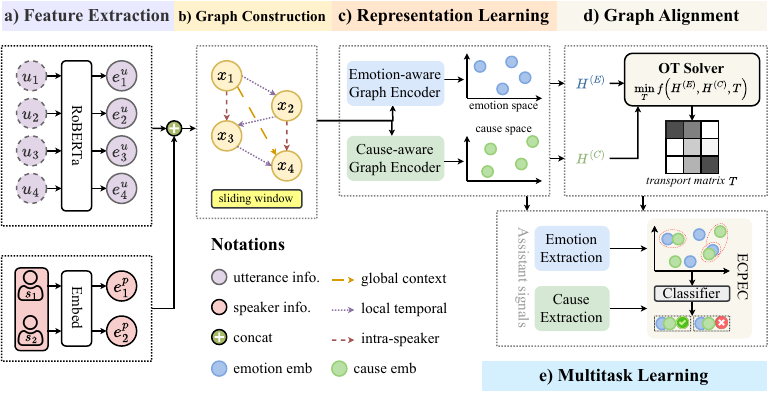}
    \caption{Overall architecture of our~\method.}
    \label{fig:archi}
\end{figure*}

Formally, given a conversation $\mathcal{C}=\{(u_1,p_{\pi(u_1)}),\cdots,(u_N,p_{\pi(u_N)})\}$ consisting of $N$ utterances, each utterance $u_j$ is associated with a speaker $p_{\pi(u_j)}$, where $\pi$ denotes a mapping from an utterance $u_i$ to the index of its corresponding speaker.
In ECPEC, the goal is to identify a set of emotion-cause pairs $\mathcal{P}=\{(u_e,u_c)\mid u_e\ \text{is caused by}\ u_c\}$ that characterizes the underlying conversational causality.
To address the limitations of existing approaches, we propose a framework termed~\method that provides a unified solution for the ECPEC task.
An overview of the proposed \method is illustrated in Figure~\ref{fig:archi}.
Specifically, each conversation is first encoded into utterance-level representations and organized as a conversation graph (\S\ref{subsec:enc_graph}).
Then, \method induces two complementary semantic views of the dialogue, namely emotion-oriented and cause-oriented representations, by applying semantic-specific graph encoding mechanisms (\S\ref{subsec:dgl}).
To explicitly model the correspondence between emotional utterances and their underlying causes, \method further formulates emotion-cause inference as a global alignment problem between the emotion-side and cause-side representations, which is solved via an optimal transport framework to enable many-to-many and globally consistent matching (\S\ref{subsec:gga}).
All components are optimized under a unified learning objective, where emotion extraction and cause extraction are introduced as auxiliary supervision signals to facilitate representation learning and ultimately improve ECPEC performance (\S\ref{subsec:multitask}).


\subsection{Encode Conversation as a Graph}\label{subsec:enc_graph}

To establish relationships between utterances while capturing both inter- and intra-speaker dependencies~\cite{ghosal2019,li2023,gao2023}, we represent each conversation as a graph $\mathcal{G}=(\mathcal{V},\mathcal{E},\bm{A})$.
Each node $v_i \in \mathcal{V}$ corresponds to an utterance $u_i$, edges $e_{ij}\in\mathcal{E}$ are constructed according to three types of dependencies, and $\bm{A}\in\mathbb{R}^{N \times N}$ is the adjacency matrix. 

\paragraph{Nodes.}
Each utterance $u_i$ is represented as a node $v_i$, initialized with a contextual embedding $\bm{x}_i^u\in\mathbb{R}^{d_u}$ obtained from a pretrained RoBERTa model~\cite{roberta}.
To incorporate speaker information, the corresponding speaker $p_{\pi(u_i)}$ is first represented as a one-hot vector and then projected into a speaker embedding $\bm{x}_{\pi(u_i)}^s\in\mathbb{R}^{d_s}$ through a learnable embedding layer.
The final speaker-aware utterance-level feature is defined as:
\begin{equation}
    \bm{x}_i = \bm{x}_i^u \oplus \bm{x}_{\pi(u_i)}^s,
\end{equation}
where $\bm{x}_i\in\mathbb{R}^{d_h}$ is the representation of node $v_i$, $\oplus$ denotes the concatenation operation.


\paragraph{Edges.}
We establish three types of relationships between utterance nodes to capture both global and local contextual dependencies in the conversation graph $\mathcal{G}$.
The global contextual edge $(v_i,v_j)$ captures long-range semantic dependencies between utterance nodes $v_i$ and $v_j$.
Such an edge exists if $\cos(\bm{x}_i, \bm{x}_j) + 1 > \tau_{\mathrm{s}}$, where $\bm{x}_i$ and $\bm{x}_j$ denote the node features of $v_i$ and $v_j$, respectively, and $\tau_{\mathrm{s}}$ is a similarity threshold.
The local contextual edge models short-range emotional dynamics.
An edge $(v_i,v_j)$ exists if $|i - j| \leq W$, where $W$ denotes the size of the sliding temporal window.
The intra-speaker edge captures speaker-specific emotional consistency.
We connect nodes $v_i$ and $v_j$ if $\pi(u_i) = \pi(u_j)$ and $i \neq j$.

\paragraph{Edge weights.}
We initialize edge weights according to the corresponding dependency type.
For global contextual edges, we initialize weights based on utterance-level semantic similarity:
\begin{equation}
e_{ij} = \frac{\cos(\bm{x}_i, \bm{x}_j) + 1}{2}.
\end{equation}
For local temporal edges, weights decay exponentially with conversational distance:
\begin{equation}
    e_{ij} = \exp(-|i-j| / \tau_\mathrm{e}),
\end{equation}
where $\tau_\mathrm{e}$ controls temporal sensitivity.
For intra-speaker edges, weights reflect speaker-specific emotional consistency across turns:
\begin{equation}
e_{ij} = \frac{\exp(-|i-j| / \tau_\mathrm{e}) + 1}{2}.
\end{equation}
All initialized edge weights are integrated into a single weighted adjacency matrix $\bm{A}$, which is treated as learnable and jointly optimized with other model parameters.



\subsection{Graph Representation Learning}\label{subsec:dgl}


Given the conversation graph $\mathcal{G}=(\mathcal{V},\mathcal{E},\bm{A})$ constructed for emotion-side and cause-side modeling, respectively, along with the same feature matrix \(\mathbf{X}=[\mathbf{x}_1,\dots,\mathbf{x}_N]^\top\in\mathbb{R}^{N\times d_h}\), we employ two independent graph encoders, $\mathrm{GNN}_E$ and $\mathrm{GNN}_C$, to learn two sets of node representations in different semantic spaces:
\begin{equation}
    \bm{H}^{(S)} = \mathrm{GNN}_S\left(\mathcal{G}, \bm{X}\right),
\end{equation}
where $S \in \{E, C\}$ denotes the emotion or cause semantic space, $\bm{H}^{(E)} \in \mathbb{R}^{N \times d_h}$ and $\bm{H}^{(C)} \in \mathbb{R}^{N \times d_h}$ denote the resulting emotion-aware and cause-aware node representations.
Following \citet{veličković2018graphattentionnetworks}, at each encoder layer, node representations $\bm{h}^{(S)} \in \bm{H}^{(S)}$ are updated via attention-weighted message passing:
\begin{equation}
\begin{aligned}
    \psi^{(S)}_{ij} &= \phi^{(S)}\!\left(\bm{h}_i^{(S)}, \bm{h}_j^{(S)}\right) A_{ij}, \\
    \alpha_{ij}^{(S)} &= \mathrm{softmax}_{j \in \mathcal{N}(i)}\!\left( \psi^{(S)}_{ij} \right), \\
    \bm{h}_i^{(S)} &= \sum_{j \in \mathcal{N}(i)} \alpha_{ij}^{(S)} \, \bm{W}^{(S)} \bm{h}_j^{(S)}, \\
\end{aligned}
\end{equation}
where $\mathcal{N}(i)$ denotes the neighborhood of node $v_i$ in $\mathcal{G}$,
$\alpha_{ij}^{(S)}$ is the normalized attention coefficient, $\bm{W}^{(S)} \in \mathbb{R}^{d_h \times d_h}$ is a learnable matrix, and $\phi^{(S)}(\cdot)$ is a learnable scoring function.
Importantly, the attention coefficient $\alpha_{ij}^{(S)}$ learned by each encoder can be viewed as a semantic-specific edge weight between $v_i$ and $v_j$ over the shared graph.
Specifically, we define a task-specific weighted adjacency matrix
$\bm{A}^{(S)} \in \mathbb{R}^{N \times N}$ as
\begin{equation}
    A^{(S)}_{ij} = \alpha^{(S)}_{ij},
\end{equation}
which captures the relative importance of edge $(v_i, v_j)$ under semantic space $S$.
As a result, the emotion-aware and cause-aware encoders implicitly induce two refined adjacency structures, denoted as $\bm{A}^{(E)}$ and $\bm{A}^{(C)}$, respectively.

 \subsection{Graph Alignment via Optimal Transport}\label{subsec:gga}
 
To capture the many-to-many relations inherent in emotion-cause pair extraction, we formulate ECPEC as a global alignment problem between emotion-aware and cause-aware representations.
Formally, given the emotion representations $\bm{H}^{(E)}=\{\bm{h}_i^{(E)}\}_{i=1}^N$ and the cause representations $\bm{H}^{(C)}=\{\bm{h}_i^{(C)}\}_{i=1}^N$, our goal is to learn a transport plan $\bm{T}\in\mathbb{R}^{N\times N}$, where each entry $T_{ij}\ge0$ indicates the soft correspondence strength between the $i$-th emotion representation $\bm{h}_i^{(E)}$ and the $j$-th cause representation $\bm{h}_j^{(C)}$.
Unlike independent pairwise scoring, the alignment is learned globally over the entire matrix $\bm{T}$, such that multiple causes can be jointly associated with the same emotion and the correspondences across different emotion-cause pairs are mutually constrained.
The alignment matrix $\bm{T}$ can be directly interpreted as a soft emotion–cause pairing matrix, where $\bm{T_{ij}}$ measures the association strength between emotion utterance $i$ and cause utterance $j$.

\paragraph{Alignment objective.}
We seek to learn the alignment matrix $\bm{T}$ by minimizing the following objective with respect to $\bm{T}$:
\begin{equation}
\begin{aligned}
    \min_{\bm{T}\ge0} \mathcal{L}(\bm{T}) &= \alpha\,\langle \bm{C}_{\text{attr}}, \bm{T}\rangle \\
        & \quad + (1-\alpha)\mathcal{L}_{\text{struct}}(\bm{T})
\end{aligned}
\end{equation}
where $\bm{C}_{\text{attr}}\in\mathbb{R}^{N\times N}$ denotes the attribute-level cost matrix, $\mathcal{L}_{\text{struct}}(\bm{T})$ denotes a structure consistency term.
$\bm{C}_{\text{attr}}$ measures semantic compatibility between emotion representation $\bm{h}_i^{(E)}\in\bm{H}^{(E)}$ and cause representation $\bm{h}_j^{(C)}\in\bm{H}^{(C)}$, with each entry defined as: 
\begin{equation}
    \bm{C}_{\text{attr}}(i,j) = 1 - \cos\left(\bm{h}_i^{(E)}, \bm{h}_j^{(C)}\right),
\end{equation}
where smaller values indicate higher semantic affinity.
The structure-level term $\mathcal{L}_{\text{struct}}(\bm{T})$ is defined as a structure consistency loss that measures the discrepancy between relational patterns encoded in the emotion-side and cause-side dialogue graphs.
Specifically, it is formulated as:
\begin{equation}
    \mathcal{L}_{\text{struct}}(\bm{T}) = \sum_{i,k,j,l} \left| A^{(E)}_{ik} - A^{(C)}_{jl}\right|^2 T_{ij} T_{kl}.
\end{equation}
Therefore, minimizing $\mathcal{L}(\bm{T})$ corresponds to finding emotion–cause pairs that are both semantically reasonable and structurally coherent within the dialogue.

\paragraph{Optimization.}
The resulting objective $\mathcal{L}(\bm{T})$ is non-linear due to the quadratic structure consistency term $\mathcal{L}_{\text{struct}}(\bm{T})$.
To efficiently minimize it in a differentiable manner, we adopt an entropy-regularized Sinkhorn scheme \citep{zeng2024} based on the standard fused Gromov-Wasserstein optimization strategy \citep{tang-etal-2023-fused}.
Starting from a uniform initialization $\bm{T}^{(0)}$, we iteratively linearize the structure consistency term around the current solution and solve a sequence of entropic optimal transport subproblems.
At iteration $t$, the linearized approximation of $\mathcal{L}_{\text{struct}}(\bm{T})$ induces an effective structure-aware cost matrix:
\begin{equation}
  \bm{C}_{\text{struct}}^{(t)}(i,j)=
  \sum_{k,l}
  \left|\bm{A}^{(E)}_{ik}-\bm{A}^{(C)}_{jl}\right|^2\,T^{(t)}_{kl},
\end{equation}
which leads to the following cost matrix used to construct a linear surrogate of $\mathcal{L}(\bm{T})$:
\begin{equation}
  \bm{C}^{(t)} = \alpha\bm{C}_{\text{attr}} + (1-\alpha)\bm{C}_{\text{struct}}^{(t)}.
\end{equation}
The updated alignment matrix $\bm{T}$ is iteratively updated via Sinkhorn normalization:
\begin{equation}
    \bm{T}^{(t+1)} =
    \mathcal{S} \left(\exp(-\bm{C}^{(t)}/\varepsilon)\right),
\end{equation}
where $\mathcal{S}$ denotes the Sinkhorn operator with standard marginal constraints, and $\varepsilon$ is the entropy regularization coefficient controlling the smoothness of the transport plan.
After convergence, we apply a row-wise softmax with temperature $\tau_{\text{r}}$ to emphasize dominant alignments:
\begin{equation}
    \widetilde{\bm{T}}=\mathrm{softmax}\!\left(\bm{T}/\tau_{\text{r}}\right).
\end{equation}
The resulting $\widetilde{\bm{T}}$ can be interpreted as a normalized alignment distribution, which reflects potential many-to-many associations between emotions and causes within each dialogue.

\subsection{Multitask Joint Learning}\label{subsec:multitask}

\method adopts a multitask joint learning framework, where all task-specific objectives are optimized simultaneously with a shared encoder and task-specific prediction heads.

\paragraph{Emotion-Cause Pair Prediction.}
Given the row-normalized alignment matrix $\widetilde{\mathbf{T}}$, each entry $\widetilde{T}_{ij}$ represents the global correspondence strength between an emotion utterance $u_i$ and a candidate cause utterance $u_j$.
To incorporate local discriminative evidence, we compute a pairwise matching score by applying a lightweight classifier to the concatenated emotion-aware and cause-aware representations of each utterance pair:
\begin{equation}
    s_{ij} = \mathcal{F}_{\mathrm{ECPEC}} \left(\left[\bm{H}^{(E)}_i, \bm{H}^{(C)}_j\right]\right),
    \label{eq:pair-score}
\end{equation}
where $\mathcal{F}_{\text{ECPEC}}$ is implemented as a lightweight MLP with a sigmoid output layer.
The final prediction score for emotion-cause pairs is obtained by combining global alignment and local evidence:
\begin{equation}
    \widehat{y}_{ij}^{(\mathrm{ECPEC})} = \beta \widetilde{T}_{ij} + (1-\beta)\,s_{ij},
\end{equation}
where $\beta$ controls the relative contribution of global correspondence and local evidence.

\paragraph{Emotion and Cause Extraction.}
We perform utterance-level emotion extraction (EE) and cause extraction (CE) using two parallel lightweight classifiers:
\begin{equation}
\begin{aligned}
    \widehat{\bm{Y}}^{(E)} &= \mathcal{F}_{\mathrm{EE}}(\bm{H}^{(E)}), \\
    \widehat{\bm{Y}}^{(C)} &= \mathcal{F}_{\mathrm{CE}}(\bm{H}^{(C)}),
\end{aligned}
\end{equation}
where $\mathcal{F}_{\mathrm{EE}}$ and $\mathcal{F}_{\mathrm{CE}}$ are implemented as lightweight MLPs with softmax output layers.


\paragraph{Joint Optimization.}
The learning of~\method is performed by minimizing $\mathcal{L}$:
\begin{equation}
    \begin{aligned}
        & \mathcal{L} = \mathcal{L}_\mathrm{ECPEC} + \lambda_{EE}\mathcal{L}_\mathrm{EE} + \lambda_{CE}\mathcal{L}_\mathrm{CE}, \\
        & \mathcal{L}_{\mathrm{EE}} = \mathrm{CE} (\widehat{\bm{Y}}^{(E)},\bm{Y}^{(E)}), \\
        & \mathcal{L}_{\mathrm{CE}} = \mathrm{CE} (\widehat{\bm{Y}}^{(C)},\bm{Y}^{(C)}),
    \end{aligned}
\end{equation}
where $\mathcal{L}_{\mathrm{EE}}$ and $\mathcal{L}_{\mathrm{CE}}$ are cross entropy losses, $\lambda_{EE}$ and $\lambda_{CE}$ controlling their contributions.
The ECPEC loss $\mathcal{L}_\mathrm{ECPEC}$ is defined as:
\begin{equation}
    \mathcal{L}_{\mathrm{ECPEC}} = \mathcal{L}_{\mathrm{pair}} + \lambda_\mathrm{OT}\,\mathcal{L}_{\mathrm{OT}}.
\end{equation}
The pair-level supervision term is computed with binary cross-entropy over the predicted pair scores:
\begin{equation}
    \mathcal{L}_{\mathrm{pair}} = \mathrm{BCE}\left( \widehat{y}^{(\mathrm{ECPEC})}, y \right),
\end{equation}
where $y\in\{0,1\}$ is the ground-truth label for the emotion-cause pair.
The OT consistency regularizer encourages the local pair-wise prediction to match the OT-derived alignment score:
\begin{equation}
    \mathcal{L}_{\text{OT}} = D_{\mathrm{KL}}\!\left( \mathrm{Bern}(s_{ij}) \| \mathrm{Bern}(\widetilde{T}_{ij}) \right).
\end{equation}
Here $D_{\mathrm{KL}}$ denotes the Kullback--Leibler divergence, $\mathrm{Bern}$ is a Bernoulli distribution, $s_{ij}$ is the local pair-wise prediction score in Equation~\eqref{eq:pair-score}, and $\tilde{T}_{ij}\in\bm{T}$ denotes the corresponding OT-derived alignment score.

\section{Experiments}
To comprehensively evaluate the proposed~\method, we formulate the following \textit{Research Questions} to guide our experiments: \\
\noindent\textit{RQ1}: How does \method compare with existing state-of-the-art approaches on ECPEC datasets? \\
\noindent\textit{RQ2}: How robust is \method in handling multi-cause scenarios compared to prior methods? \\
\noindent\textit{RQ3}: How do key components of \method contribute to ECPEC performance? \\
\noindent\textit{RQ4}: Can \method provide interpretable emotion--cause alignments and meaningful insights through qualitative analysis? \\

\subsection{Experimental Setups}

\paragraph{Datasets.}
We evaluate our method on three representative datasets, described below.
RECCON~\citep{poria2021} is a widely used benchmark that consists of two subsets: \textbf{RECCON-DD}, annotated from DailyDialog~\citep{li2017}, serves as the main corpus for model training and evaluation, while \textbf{RECCON-IE}, annotated from IEMOCAP~\citep{busso2008}, is a smaller subset used exclusively to test the generalization ability. \textbf{ECF}~\citep{wang2023} is a multimodal benchmark derived from the sitcom \textit{Friends}, providing annotated emotion–cause pairs across text, audio, and visual modalities, with many emotions triggered by multiple utterances.
The dataset statistics are illustrated in Table~\ref{tab:dataset}.

\begin{table}[!t]
    \centering
    \resizebox{.49\textwidth}{!}{
    \begin{tabular}{lrrrr}
        \toprule
        \textbf{Dataset} & \textbf{\#Dlg.} & \textbf{\#Utt.} & \textbf{\#Pairs} & \textbf{Partition} \\
        \midrule
        RECCON-DD  & 1,106 & 11,104 & 5,861 & 75/5/20 \\
        RECCON-IE  & 16 & 665 & 1154 & test only \\
        ECF & 1,374 & 13,619 & 9,794 & 70/10/20 \\
        \bottomrule
\end{tabular}}
\caption{Dataset statistics.}
\label{tab:dataset}
\end{table}

\paragraph{Baselines.}
We compare our method against seven representative baselines, which can be broadly grouped into three categories: \\
\textit{1) Method based on general pre-trained model:}
Following~\citet{poria2021}, we adopt a pretrained RoBERTa~\cite{roberta} model with a classification layer to identify emotion–cause pairs, which we refer to as \textbf{RECCON}, serving as a benchmark baseline.\\
\textit{2) Methods with sequential modeling:}
\textbf{MECPE-2steps}~\citep{wang2023} adopts a two-step pipeline that first extracts candidate emotions and causes sets by a shared BiLSTM and then filters valid pairs with another BiLSTM.
\textbf{PRG-MoE}~\citep{jeong2023} constructs relational graph with a mixture-of-experts, where a gating network aggregates diverse relational patterns.
\textbf{Joint-Xatt}~\citep{li2023} utilize cross-attention to model emotion–cause dependencies. \\
\textit{3) Methods based on graph modeling:}
\textbf{Joint-GCN}~\citep{li2023} extends Joint-Xatt by replacing cross-attention with graph convolutional network to model inter-utterance relations.
\textbf{MRC}~\citep{liu2023} reformulates ECPEC as machine reading comprehension and employs GNNs to encode dialogue structure.
\textbf{CENTER}~\citep{wang2024} builds a center event–aware graph with contrastive objectives for pair-level discrimination.
\textbf{MultiCauseNet}~\citep{ma2025} leverages a multimodal graph and employs Graph Attention Networks with temporal attention to prioritize relevant features across text, audio, and video modalities. \\
\textit{4) Methods based on generative frameworks and LLMs:}
\textbf{GMEC}~\citep{ju2025} transforms the extraction task into a generative question-answering paradigm, utilizing Large Language Models as implicit knowledge engines to capture both linguistic and visual cues.

\paragraph{Metrics.} 
Following previous work~\citep{xia2019}, we adopt F1-score (F1), Precision (P), and Recall (R) as evaluation metrics.

\paragraph{Implementation Details.}
We derive textual features for each utterance using a pretrained RoBERTa model.
Unless otherwise specified, hyperparameters are set as follows:
the temporal window size $W=5$,
utterance and speaker embedding dimensions $d_u=768$ and $d_s=50$,
similarity and decay parameters $\tau_{\mathrm{s}}=0.5$, $\tau_{\mathrm{e}}=2.0$, and $\tau_{\mathrm{r}}=1.0$,
and weighting coefficients $\alpha=0.8$, $\beta=0.4$, $\varepsilon=0.5$, $\lambda_\mathrm{EE}=0.2$, $\lambda_\mathrm{CE}=0.4$, and $\lambda_\mathrm{OT}=1.0$.
Model training is conducted using the AdamW optimizer~\cite{adamw} with a learning rate of $1e-4$.
We employ a ReduceLROnPlateau scheduler and early stopping based on validation performance.
All experiments are implemented in PyTorch~\cite{pytorch} and executed on a single NVIDIA RTX 4090 GPU.
For a consistent comparison, all baselines are reimplemented based on their publicly released code or original descriptions, and trained under the same experimental settings.

\begin{table*}[!t]
    \centering
    \small
    \renewcommand{\arraystretch}{1.1}
    \setlength{\tabcolsep}{7pt}

    \begin{tabular}{
        l
        *{9}{S[table-format=2.2, detect-weight=true, detect-inline-weight=math]}
    }
        \toprule
        \multirow{2}{*}{\textbf{Method}}
        & \multicolumn{3}{c}{\textbf{RECCON-DD}} 
        & \multicolumn{3}{c}{\textbf{RECCON-IE}} 
        & \multicolumn{3}{c}{\textbf{ECF}} \\
        \cmidrule(lr){2-4}\cmidrule(lr){5-7}\cmidrule(lr){8-10}
        & P & R & F1
        & P & R & F1
        & P & R & F1 \\
        
        \midrule
        RECCON       & 49.31 & 33.19 & 39.68 & 51.04 & 11.00 & 18.10 & 30.26 & 37.58 & 33.52 \\
        MECPE-2steps & 49.34 & 47.37 & 48.34 & 27.31 & 6.30  & 10.24 & \underline{57.64} & 48.72 & 52.71 \\
        PRG-MoE      & \textbf{58.95} & \underline{55.67} & \underline{57.26} 
                     & \underline{51.95} & 20.02 & \underline{28.90} 
                     & 47.11 & 55.27 & 50.86 \\
        Joint-Xatt   & 28.64 & 40.43 & 33.53 & 28.37 & 12.30 & 17.16 & 42.65 & 39.19 & 40.85 \\
        Joint-GCN    & 30.79 & 36.88 & 33.56 & 27.49 & 16.67 & 20.75 & 40.29 & 42.33 & 41.28 \\
        MRC          & 52.19 & 52.86 & 52.47 & \textbf{59.59} & 16.08 & 20.96 & 44.46 & 57.65 & 50.20 \\
        CENTER       & 47.39 & 46.88 & 47.13 & 34.92 & \underline{24.38} & 28.71 & 47.90 & 43.65 & 44.75 \\
        MultiCauseNet & \multicolumn{1}{c}{-} & \multicolumn{1}{c}{-} & 
\multicolumn{1}{c}{-} & \multicolumn{1}{c}{-} & \multicolumn{1}{c}{-} & 
\multicolumn{1}{c}{-} & 53.27 & \underline{59.10} & \underline{55.12} \\
        GMEC         & 55.97 & 50.46 & 53.07 & 46.79 & 20.24 & 28.25 & \textbf{60.41} & 50.03 & 54.73 \\
        
        \midrule
        \textbf{Ours}  
                     & \underline{56.31} & \textbf{61.60} & \textbf{58.83}
                     & 42.54 & \textbf{29.29} & \textbf{34.69}
                     & 55.01 & \textbf{60.67} & \textbf{57.70} \\
        \bottomrule
    \end{tabular}
    \caption{Comparison results of ECPEC task.} 
    \label{tab:main_result}
\end{table*}

\subsection{Results and Discussion}

\subsubsection{Overall Performance (RQ1)}

We evaluate all methods on the ECPEC task across three benchmarks, namely RECCON-DD, RECCON-IE, and ECF.
As reported in Table~\ref{tab:main_result}, \method consistently achieves the highest recall and F1-score on all three datasets.
On RECCON-DD, \method attains an F1 score of 58.83, outperforming the strongest baseline in terms of F1, PRG-MoE (57.26), by a relative improvement of $+2.7\%$.
On the smaller and cross-domain RECCON-IE dataset, \method achieves the best F1 score of 34.69, surpassing the strongest baseline (28.90) by a substantial relative gain of $+20.0\%$, which highlights its strong generalization capability.
Similarly, on ECF, \method delivers a notable relative improvement of $+9.5\%$ over MECPE-2steps (52.71) in terms of F1-score.
We also observe that \method does not achieve the highest precision on most datasets. 
This behavior is consistent with the design of the soft optimal transport alignment, which encourages broader semantic matching between emotion and cause representations and therefore favors higher recall at the potential expense of precision.

\begin{table}[!t]
    \centering
    \footnotesize
    \setlength{\tabcolsep}{3.2pt}
    \begin{tabular}{lccc}
        \toprule
        \textbf{Method} & \textbf{RECCON-DD} & \textbf{RECCON-IE} & \textbf{ECF} \\
        \midrule
        RECCON         & 28.27 & ~~8.76  & 23.75 \\
        MECPE-2steps   & 33.61 &  11.48  & 28.71 \\
        PRG-MoE        & {37.84} &  {21.62}  & {33.71} \\
        Joint-Xatt     & 23.18 & ~~4.62  & 25.15 \\
        Joint-GCN      & 23.73 & ~~5.08  & 26.96 \\
        MRC            & 33.39 & ~~3.22  & 25.64 \\
        CENTER         & 34.09 & ~~9.73  & 28.20 \\
        \midrule
        \textbf{\method} & \textbf{38.33} & \textbf{25.33} & \textbf{35.55} \\
        \bottomrule
    \end{tabular}
    \caption{Comparison of F1-scores on the multi-cause scenario.}
    \label{tab:multi_cause_result}
\end{table}


\subsubsection{Multi-Cause Study (RQ2)}

As mentioned above, multi-cause scenarios introduce additional challenges for ECPEC.
To evaluate model robustness under such settings, we therefore construct three multi-cause test subsets from RECCON-DD, RECCON-IE, and ECF, by selecting all dialogues in the original test splits where a single target emotion is annotated with two or more distinct causes.
We then re-evaluate all models on these subsets to assess their robustness in capturing multiple causal triggers.
As shown in Table~\ref{tab:multi_cause_result}, all models suffer from a notable performance drop, confirming the inherent difficulty of multi-cause prediction.
Nevertheless, \method consistently achieves the best F1 scores across all three subsets.
We attribute these improvements to the global alignment formulation in \method, which models emotion–cause relations as soft many-to-many correspondences between emotion-oriented and cause-oriented representations, enabling joint reasoning over dispersed causal evidence for each emotion.

\begin{table}[!t]
    \centering
    \footnotesize
    \setlength{\tabcolsep}{3.2pt}
    \begin{tabular}{lccc}
        \toprule
         & \textbf{RECCON-DD} & \textbf{RECCON-IE} & \textbf{ECF} \\
        \midrule
        Full model & \bfseries 58.83 & \bfseries 34.69 & \bfseries 57.70 \\
        \midrule
        \quad w/o SRL &  56.72 &  31.22 &  55.77 \\
        \quad w/o GA  &  55.26 &  29.08 & 53.82 \\
        \quad w/o SRL \& GA & 53.66 & 28.18 & 52.54\\
        \midrule
        \quad w/o EE & 58.44 & 34.43 & 57.43\\
        \quad w/o CE & 57.99 & 34.07 & 57.11\\
        \quad w/o CE \& EE & 57.15 & 33.57 & 56.54\\
        \bottomrule
    \end{tabular}
    \caption{Ablation study.}
    \label{tab:ablation}
\end{table}

\subsubsection{Ablation Study (RQ3)}
To verify the effectiveness of the key design principles in \method, we conduct ablation experiments by removing separated representation learning (SRL; see $\S$\ref{subsec:dgl}), global alignment (GA; see $\S$\ref{subsec:gga}), and auxiliary supervision (i.e., EE and CE).
Specifically, w/o SRL collapses emotion-oriented and cause-oriented encoders into a single graph encoder that learns a shared representation for all utterances, while w/o GA removes the alignment module and performs emotion--cause prediction based solely on independent pairwise scores.
As shown in Table~\ref{tab:ablation}, removing either SRL or GA consistently degrades ECPEC F1 performance across all datasets, with a more pronounced drop observed when GA is disabled, underscoring the importance of soft many-to-many alignment for modeling complex emotion-cause relations.
We further observe that auxiliary supervision is beneficial to ECPEC.
Removing EE or CE individually leads to mild performance drops, while jointly removing both results in a more noticeable degradation, indicating that EE and CE provide complementary support for representation learning.

\subsubsection{Qualitative Analysis (RQ4)}

\begin{figure}[!t]
    \centering
    \begin{subfigure}{\linewidth}
        \centering
        \includegraphics[width=\linewidth]{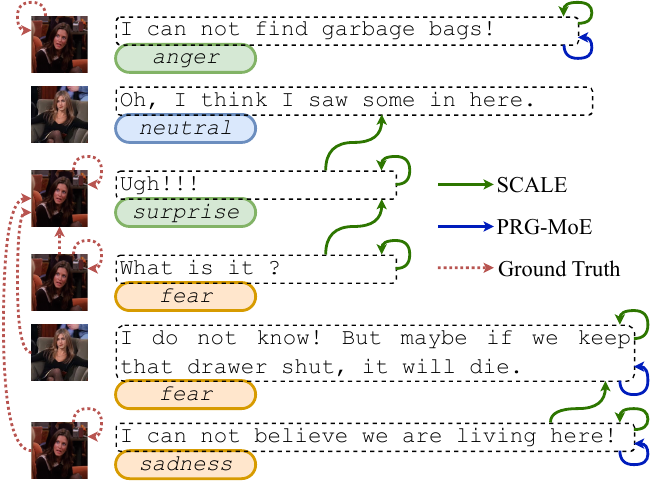}
        \caption{Qualitative comparison.}
        \label{fig:case_study}
    \end{subfigure}

    \vspace{0.5em}

    \begin{subfigure}{0.85\linewidth}
        \centering
        \includegraphics[width=\linewidth]{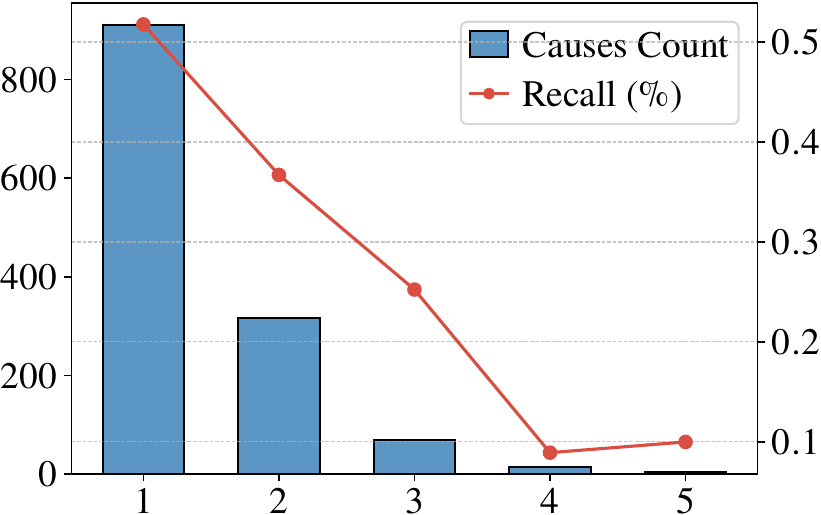}
        \caption{Cause-number distribution and corresponding recall on ECF dataset.}
        \label{fig:error}
    \end{subfigure}
    \caption{Qualitative analysis.}
\end{figure}

\paragraph{Case Study}
Figure~\ref{fig:case_study} presents a qualitative comparison between \method and the strongest baseline PRG-MoE on a dialogue sampled from the ECF dataset.
Among the seven ground-truth emotion--cause pairs, \method correctly predicts five, whereas PRG-MoE identifies only two, indicating a clear performance gap.
For long-distance dependencies, such as $(u_6, u_3)$ and $(u_5, u_3)$, both models fail to recover the correct relations, suggesting that capturing causal cues separated by large conversational gaps remains challenging.
In contrast, for short-distance dependencies, including $(u_1, u_1)$, $(u_3, u_3)$, $(u_4, u_4)$, $(u_4, u_3)$, and $(u_5, u_5)$, \method achieves perfect predictions.
Regarding multi-cause scenarios, \method successfully identifies the dual-cause emotion $[(u_4, u_3), (u_4, u_4)]$, but fails to capture both causes in $[(u_6, u_6), (u_6, u_3)]$.
These observations suggest that while the global alignment mechanism enables flexible one-to-many reasoning, modeling long-range causal dependencies remains an open challenge.

\paragraph{Error Analysis.}
To further analyze the behavior of \method in multi-cause scenarios, we examine its performance on the ECF dataset.
As shown in Figure~\ref{fig:error}, most instances involve a single cause, while samples with multiple causes are increasingly scarce.
Despite achieving the best overall performance on the multi-cause subset of ECF, \method exhibits decreasing recall as the number of causes increases, since recall requires all causes associated with an emotion to be correctly identified, indicating that exhaustive cause retrieval in complex multi-cause settings remains challenging.

\subsubsection{Auxiliary Analysis}

\paragraph{Performance on EE and CE.}
To provide additional context on the intermediate subtasks, we evaluate the performance of \method on emotion extraction (EE) and cause extraction (CE).
As shown in Table~\ref{tab:all-ee-ce}, \method yields reasonable performance on both EE and CE across datasets, without relying on task-specific architectural designs.
It is worth noting that \method is primarily optimized for the ECPEC objective, while EE and CE are incorporated as auxiliary supervision during training rather than standalone targets.

\begin{table}[!t]
    \centering
    \footnotesize
    \setlength{\tabcolsep}{1.8pt}
    \begin{tabular}{lcccccc}
        \toprule
        \multirow{2}{*}{\textbf{Method}} &
        \multicolumn{2}{c}{\textbf{RECCON-DD}} &
        \multicolumn{2}{c}{\textbf{RECCON-IE}} &
        \multicolumn{2}{c}{\textbf{ECF}} \\
        \cmidrule(lr){2-3}\cmidrule(lr){4-5}\cmidrule(lr){6-7}
        & EE & CE & EE & CE & EE & CE \\
        \midrule
        MECPE-2steps    & 71.30 & 65.81 & 42.52 & 44.49 & \textbf{79.10} & \textbf{70.13} \\
        PRG-MoE         & 73.86 & - & 57.29 & - & 71.82 & - \\
        Joint-Xatt      & 58.71 & 51.93 & 47.34 & 40.26 & 67.75 & 62.73 \\
        Joint-GCN       & 61.87 & 51.35 & 46.58 & 35.75 & 68.31 & 64.05 \\
        MRC             & \textbf{75.49} & - & 39.54 & - & 74.08 & - \\
        CENTER          & 68.32 & - & 49.61 & - & 67.07 & - \\
        \midrule
        \textbf{\method} & 73.23 & \multicolumn{1}{c}{\textbf{67.87}} & \multicolumn{1}{c}{55.57} & \multicolumn{1}{c}{\textbf{54.90}} & \multicolumn{1}{c}{76.10} & 61.81 \\
        \bottomrule
    \end{tabular}
    \caption{Comparative results of EE and CE subtasks (F1-score) across three datasets.}
    \label{tab:all-ee-ce}
\end{table}

\paragraph{Comparison with recent LLMs.}
We compare \method with three recent LLMs,using a few-shot prompting strategy with four in-context examples and strict JSON output constraints to ensure fair evaluation. The complete prompt template and example dialogues are provided in Appendix~\ref{apx:LLMs baselines}.
As shown in Table~\ref{tab:llm_fewshot}, \method achieves higher F1-scores than the best-performing LLM on both datasets.
These results suggest that explicit modeling of conversational structure and emotion--cause relations remains advantageous for ECPEC, even in the presence of strong prompt-based LLM baselines.

\begin{table}[!t]
    \centering
    \footnotesize
    \begin{tabular}{lcc}
        \toprule
        \textbf{Method} & \multicolumn{1}{c}{\textbf{RECCON-DD}} & \multicolumn{1}{c}{\textbf{ECF}} \\
        \midrule
        DeepSeek-V3.2 & 47.11 & 42.81 \\
        \mbox{GPT-5.1~Instant} & 55.26 & 54.76 \\
        \mbox{Gemini-3-pro-preview} & 56.08 & 55.42 \\
        \midrule
        \textbf{\method} & \bfseries 58.83 & \bfseries 57.70 \\
        \bottomrule
    \end{tabular}
    \caption{Comparison with recent LLMs.}
    \label{tab:llm_fewshot}
\end{table}

\begin{figure}[!t]
    \centering
    \begin{subfigure}{0.7\linewidth}
        \centering
        \includegraphics[width=\linewidth]{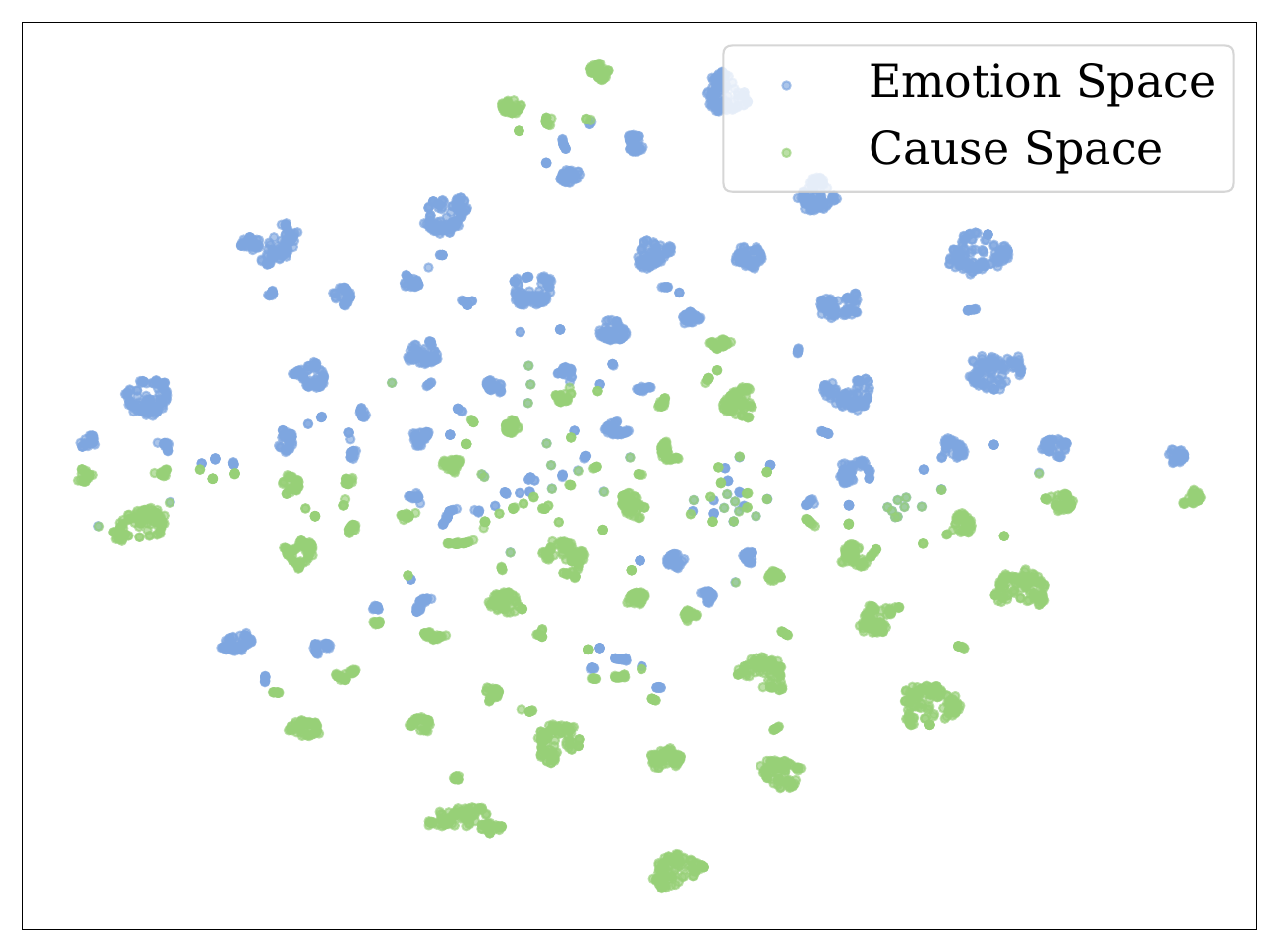}
        \caption{t-SNE visualization of \method.}
        \label{fig:tSNE_ours}
    \end{subfigure}

    \vspace{0.5em}
    \begin{subfigure}{0.7\linewidth}
        \centering
        \includegraphics[width=\linewidth]{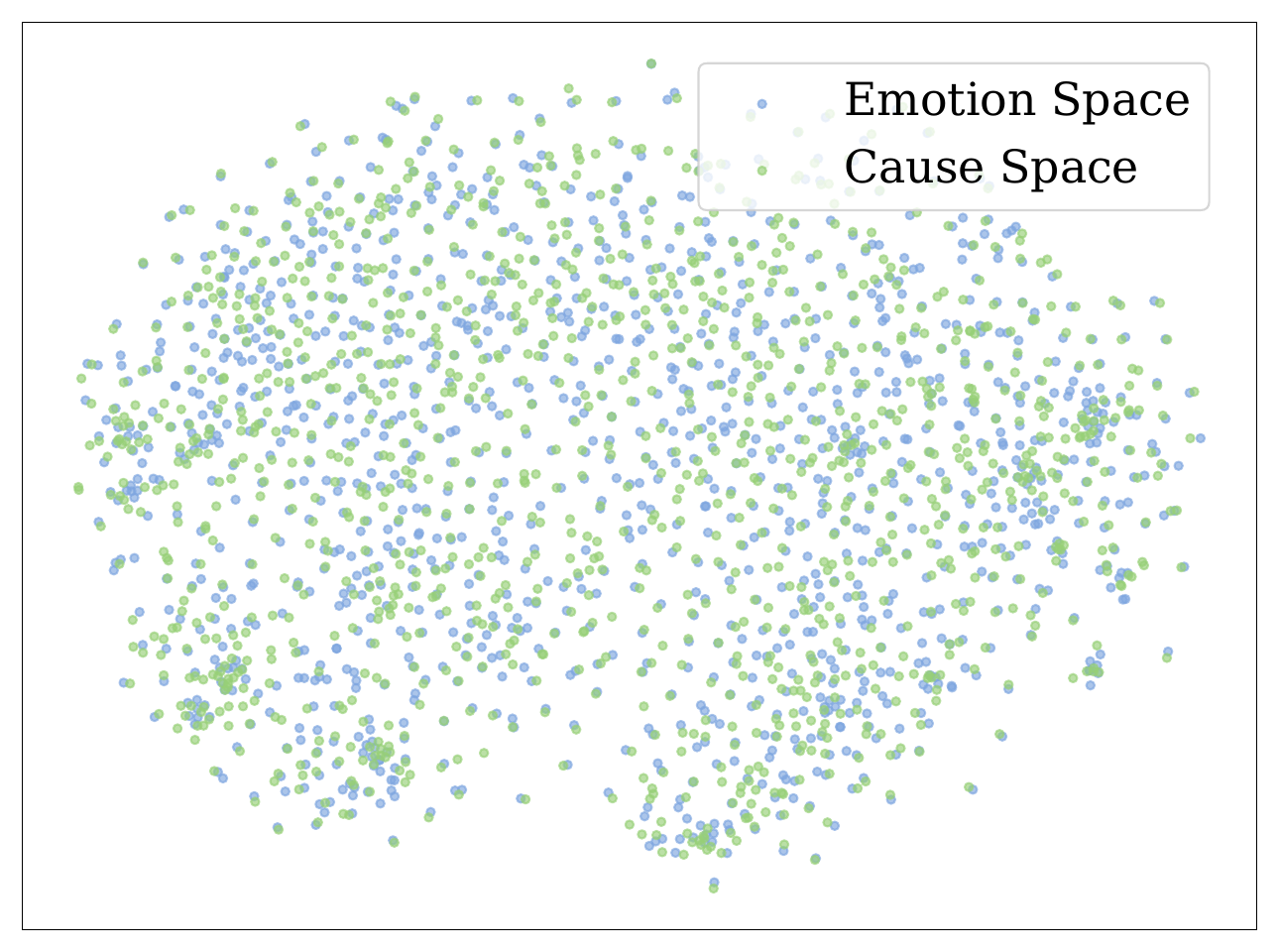}
        \caption{t-SNE visualization of MECPE-2steps.}
        \label{fig:tSNE_MECPE2steps}
    \end{subfigure}
   
    \caption{t-SNE visualization of emotion and cause representations.}
\end{figure}

\subsubsection{Visualization Analysis}
We further visualize the latent features of the last hidden layer before the classifier using t-SNE, illustrating the distributions of emotion space and cause space for MECPE-2steps and \method.
As illustrated in Figure \ref{fig:tSNE_MECPE2steps}, the node features  are uniformly intertwined, forming a dense cluster without clear decision boundaries. This phenomenon indicates a severe semantic confusion, where the model fails to explicitly distinguish whether an utterance serves as an emotion carrier or a cause trigger during the encoding stage.
In contrast, Figure \ref{fig:tSNE_ours} demonstrates that \method maps emotion and cause features into distinct, highly discriminative subspaces, validating that dual-graph module achieves effective semantic decoupling.
Furthermore, instead of forming a single massive cluster, our features spontaneously aggregate into dozens of compact, high-cohesion local clusters, suggesting that \method captures more fine-grained structural patterns.
\section{Conclusion}

In this paper, we proposed \textbf{\method}, a semantic alignment framework for emotion--cause pair extraction in conversations.
By decoupling emotion-oriented and cause-oriented semantics and modeling their interactions through global alignment, \method reformulates ECPEC as a many-to-many reasoning problem over conversational structure.
This design enables more holistic modeling of complex causal dependencies beyond independent pairwise prediction.
Extensive experiments on three benchmark datasets demonstrate that \method consistently outperforms existing approaches, particularly in challenging multi-cause scenarios, validating its effectiveness and robustness.

\section*{Limitations}

The proposed \textbf{\method} focuses on textual conversations and does not incorporate other modalities such as acoustic or visual signals.
In real-world scenarios, emotion expression and causal cues may span multiple modalities, which could provide complementary information beyond text alone.
While the alignment-based formulation of \method is in principle compatible with multimodal representations, extending the framework to fully multimodal ECPEC settings is beyond the scope of this work and remains an interesting direction for future research.

Additionally, emotion-cause relations in conversation are often ambiguous and context-dependent, and models trained on annotated datasets may inherit annotation bias or incomplete causal assumptions.
Therefore, the proposed framework should be used as an assistive tool rather than a definitive explanation of emotion causality.

\section*{Acknowledgments}
This work was supported by the National Natural Science Foundation of China (Grant No. PA2023GDGP0109) and the Research Startup Fund of Hefei University of Technology (Grant No. JZ2023HGQA0470).

\bibliography{bib/erc,bib/ecpe,bib/ecpec,bib/ot,bib/other}

\appendix

\section{Related Work}

\subsection{Emotion-Cause Pair Extraction.}
Research on emotion-cause analysis originated from the Emotion Cause Extraction (ECE) task~\citep{lee2010,gui2016,li2018}, which aims to identify the cause span corresponding to a given emotion. 
To overcome the limitation of requiring emotion annotation before cause extraction, \citet{xia2019} proposed the Emotion-Cause Pair Extraction (ECPE) task, which jointly extracts emotions and their causes, and inspired a line of subsequent studies~\citep{ding2020,chen2020,fan2021}. 
While remarkable progress has been made in ECPE research, the majority of existing approaches~\citep{fan2021,chen2022,zhu2024} are confined to document-level corpora, where emotions and their corresponding causes are expressed within a single, coherent narrative flow.
Such clause-level formulations inherently neglect the distinctive characteristics of dialogues, such as speaker role alternation, intertwined emotional events, and long-range conversational dependencies, thereby highlighting the necessity of extending ECPE into conversational contexts.

\subsection{Emotion-Cause Pair Extraction in Conversation.}
Early efforts on conversational cause analysis began with RECCON~\citep{poria2021}, which focused on cause recognition rather than emotion-cause pair extraction. 
\citet{li2023} formally introduced the ECPEC task and released the ConvECPE dataset, along with a two-step framework that explicitly models conversational properties such as context dependence and speaker interactivity.
Subsequent works explored more sophisticated modeling, such as pair-relations-guided mixture-of-experts system PRG-MOE~\citep{jeong2023}, machine reading comprehension-based method MRC~\citep{liu2023}, global-view speaker-aware frameworks GSESE~\citep{an2023}, and event-guided ECPEC frameworks CENTER~\citep{wang2024}.
Beyond text, multimodal ECPEC has been studied with the ECF dataset \citep{wang2023} and improved by cross-modality interaction mechanisms such as HiLo \citep{li2024}. 
Despite these advances, existing methods still follow the pairwise classification paradigm and thus fail to capture global many-to-many alignments between emotions and causes in dialogues, leaving robust causal modeling an open challenge.

\subsection{Graph Alignment and Optimal Transport}
Graph alignment aims to identify correspondences between nodes across related graphs~\cite{saxena2024,skitsas2023,trung2020}, a problem widely studied in network analysis and data integration. 
As a global matching problem, it can be naturally addressed by Optimal Transport (OT)~\cite{zeng2024,wang2023a,xu2019}, which computes a global coupling between two sets that minimizes transportation cost and captures many-to-many correspondences under a global optimization objective~\cite{zeng2023,maretic2019}.
This property makes OT particularly suitable for aligning emotions and causes in dialogues, where multiple candidates may coexist and local decisions can be insufficient. 
However, OT has not yet been explored in ECPEC, which motivates reformulating our task as a global graph alignment problem through the integration of dual graph learning and OT.

\section{LLMs Baselines }\label{apx:LLMs baselines}
\begin{figure*}[h]
    \centering
    \includegraphics[width=.9\textwidth]{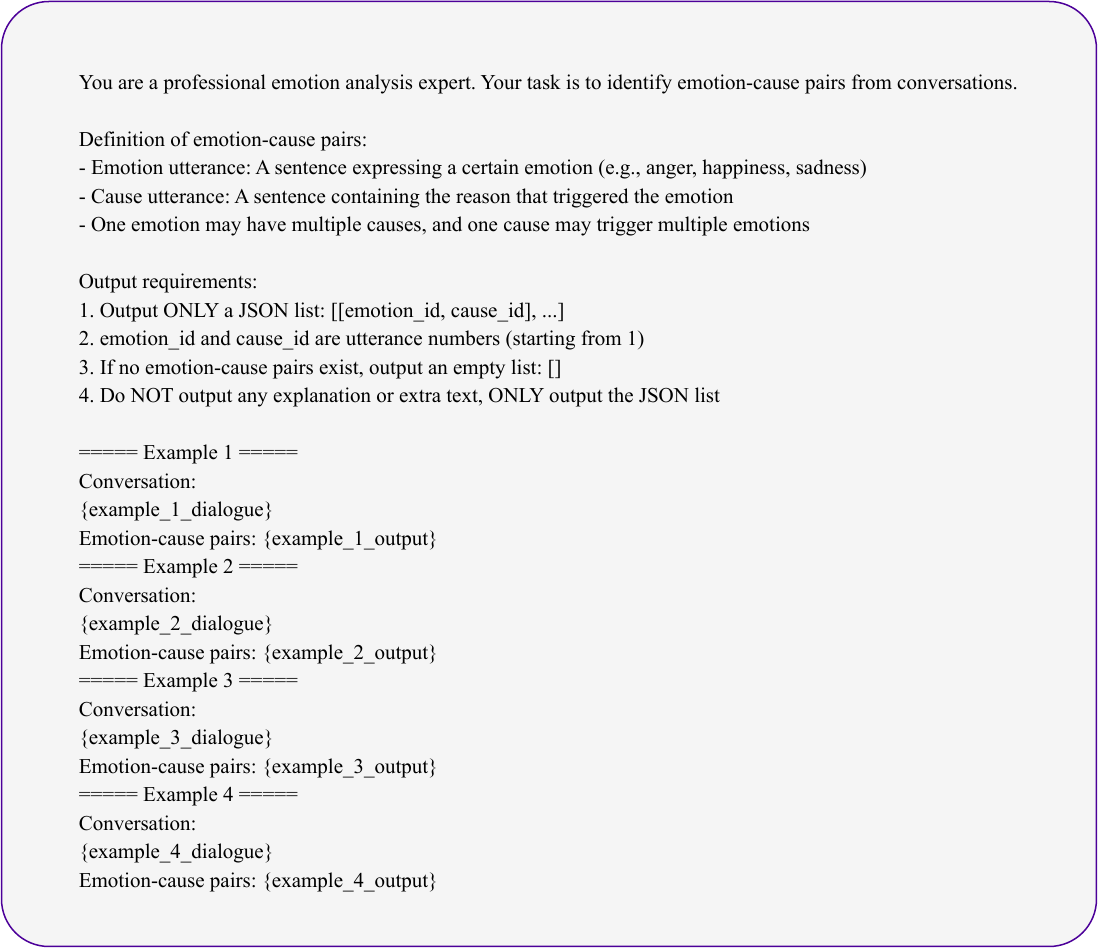}
    \caption{The system prompt template.}
    \label{fig:prompt}
\end{figure*}

To improve transparency and reproducibility, the complete prompt template is provided in Figure~\ref{fig:prompt}.

\section{Key Hyperparameters Experiments }
To further investigate the effects of key hyperparameters, we conduct additional experiments, as shown in Table~\ref{tab:hyperparameters}. The results reveal that neither attribute-only nor structure-only alignment yields optimal performance. Notably, enforcing pure alignment ($\beta = 1$) results in a significant degradation in performance. Furthermore, the model maintains stable performance over a reasonably wide range of window sizes.

\begin{table}[htbp]
    \centering
    \footnotesize
    \setlength{\tabcolsep}{10pt} 
    
    \begin{subtable}{\linewidth}
        \centering
        \begin{tabular}{lccc}
            \toprule
             & \textbf{P} & \textbf{R} & \textbf{F1} \\
            \midrule
            0.0 & 54.77 & 54.05 & 54.41 \\
            0.2 & 55.01 & 60.67 & 57.70 \\
            0.4 & 54.05 & 59.97 & 56.85 \\
            0.6 & 55.93 & 57.99 & 56.94 \\
            0.8 & 55.11 & 60.49 & 57.67 \\
            1.0 & 55.04 & 58.09 & 56.52 \\
            \bottomrule
        \end{tabular}
        \caption{Effect of $\alpha$ on performance.}
        \label{subtab:alpha_v}
    \end{subtable}
    
    \vspace{0.4cm} %
    
    \begin{subtable}{\linewidth}
        \centering
        \begin{tabular}{lccc}
            \toprule
             & \textbf{P} & \textbf{R} & \textbf{F1} \\
            \midrule
            0.0 & 56.01 & 57.04 & 56.52 \\
            0.2 & 55.53 & 58.95 & 57.18 \\
            0.4 & 55.01 & 60.67 & 57.70 \\
            0.6 & 55.39 & 57.14 & 56.25 \\
            0.8 & 56.30 & 43.41 & 49.02 \\
            1.0 & 41.25 & 10.58 & 16.84 \\
            \bottomrule
        \end{tabular}
        \caption{Effect of $\beta$ on performance.}
        \label{subtab:beta_v}
    \end{subtable}
    
    \vspace{0.4cm} 
    
    \begin{subtable}{\linewidth}
        \centering
        \begin{tabular}{lccc}
            \toprule
             & \textbf{P} & \textbf{R} & \textbf{F1} \\
            \midrule
            3  & 54.93 & 59.32 & 57.04 \\
            4  & 55.90 & 58.44 & 57.14 \\
            5  & 55.01 & 60.67 & 57.70 \\
            6  & 54.35 & 59.76 & 56.93 \\
            8  & 53.97 & 60.75 & 57.15 \\
            10 & 53.63 & 60.22 & 56.73 \\
            \bottomrule
        \end{tabular}
        \caption{Effect of $window\_size$ on performance.}
        \label{subtab:window_v}
    \end{subtable}

    \caption{Hyperparameter experiments on the ECF dataset with respect to $\alpha$, $\beta$, and window\_size.}
    \label{tab:hyperparameters}
\end{table}

\section{Computational Complexity Analysis }

Theoretically, for a conversation with $N$ utterances and $|\mathcal{E}|$ edges, one GNN layer costs $O(|\mathcal{E}|d)$. With $L$ layers and two encoders, encoding cost is $O(2L|\mathcal{E}|d)$, i.e., only a constant-factor increase. The Sinkhorn alignment operates on an $N \times N$ matrix with per-iteration cost $O(N^2)$. Since dialogue length is small ($N < 30$), the alignment remains lightweight.
Empirical results in Table~\ref{tab:efficiency} compare parameter size, FLOPs, and peak memory usage, demonstrating that SCALE achieves competitive performance with substantially lower computational cost.

\begin{table}[htbp]
    \centering
    \footnotesize
    \begin{tabular}{lccc}
        \toprule
        \textbf{Model} & \textbf{Params} & \textbf{FLOPs} & \textbf{Memory} \\
        \midrule
        PRG-MoE & 110M & $\sim$220G & 22G \\
        GMEC & 450M & $\sim$360G & 20G \\
        \midrule
        \textbf{SCALE} & \bfseries 8.2M & \bfseries 2.15G & \bfseries 7G \\
        \bottomrule
    \end{tabular}
    \caption{Comparison of Model Efficiency.}
    \label{tab:efficiency}
\end{table}

\section{Multimodal Extension }
Although~\method is not primarily designed for multimodal modeling, it can be easily extended to handle multimodal inputs. 
As the ECF dataset provides text, audio, and video modalities, we follow prior work by concatenating unimodal features as a simple fusion strategy, since multimodal fusion is not the main focus of this paper. 
As shown in Table~\ref{tab:multimodal}, ~\method achieves consistent improvements over the text-only variant when additional modalities are incorporated, demonstrating its adaptability and robustness across multimodal settings.

\begin{table}[htbp]
    \centering
    \footnotesize
    \begin{tabular}{cccc}
        \toprule
        Text & Audio & Video & F1 \\
        \midrule
        + & - & - & 57.70 \\
        + & + & - & 58.13 \\
        + & - & + & 58.07 \\
        + & + & + & \textbf{58.63} \\
        \bottomrule
    \end{tabular}
    \caption{Multimodal evaluation of~\method on the ECF dataset.}
    \label{tab:multimodal}
\end{table}

\end{document}